\RequirePackage{fix-cm}
\documentclass[twocolumn]{svjour3}          
\smartqed  
\usepackage{graphicx}
\usepackage{amsfonts}
\usepackage{amsmath}
\usepackage{color}
\begin{document}

\title{Boosting Scene Character Recognition by Learning Canonical Forms of Glyphs}

\author{Yizhi Wang         \and
        Zhouhui Lian      \and
        Yingmin Tang       \and
        Jianguo Xiao
}


\institute{Y. Wang \and Z. Lian \and Y. Tang \and J. Xiao \at
              Institute of Computer Science and Technology, Peking University\\
             \email{\{wangyizhi,lianzhouhui,tangyingmin,xjg\}@pku.edu.cn}
              Tel.: +86-10-82529245\\
}
\date{Received: date / Accepted: date}
\maketitle

\begin{abstract}
As one of the fundamental problems in document analysis, scene character recognition has attracted considerable interests in recent years.
But the problem is still considered to be extremely challenging due to many uncontrollable factors including glyph transformation, blur, noisy background, uneven illumination, etc.
In this paper, we propose a novel methodology for boosting scene character recognition by learning canonical forms of glyphs, based on the fact that characters appearing in scene images are all derived from their corresponding canonical forms.
Our key observation is that more discriminative features can be learned by solving specially-designed generative tasks compared to traditional classification-based feature learning frameworks.
Specifically, we design a GAN-based model to make the learned deep feature of a given scene character be capable of reconstructing corresponding glyphs in a number of standard font styles. In this manner, we obtain deep features for scene characters that are more discriminative in recognition and less sensitive against the above-mentioned factors.
Our experiments conducted on several publicly-available databases demonstrate the superiority of our method compared to the state of the art.
\keywords{Character recognition \and Canonical forms of glyphs \and Multi-task learning \and Deep learning}
\end{abstract}

\section{Introduction}
Scene character recognition (SCR) is an important and challenging problem in areas of Document Analysis, Pattern Recognition and Computer Vision.
Automatic reading of characters in natural scenes is very useful to a wide range of applications such as autonomous vehicle navigation, textual image retrieval, machine translation, etc. \par
There are two widely used feature representations for scene character recognition including hand-crafted features and neural network based features.
The majority of hand-crafted feature based methods employ HOG (Histogram of Oriented Gradient)~\cite{Dalal2005Histograms} like features for character recognition, such as~\cite{Tian2016Multilingual,zhang2016natural}.
However, these traditional methods based on handcrafted features are not able to satisfactorily deal with noisy data in natural scenes.
Recently, deep learning based models have been presented for solving character recognition problem.
Convolutional Neural Networks (CNNs), which possess a powerful feature expression ability, are utilized to recognize characters in many works, such as~\cite{Wang2013End,Bissacco2013PhotoOCR,Jaderberg2014Deep,Wang2017Learning}.

\begin{figure}[t!]
  \centering
  \includegraphics[width=8cm]{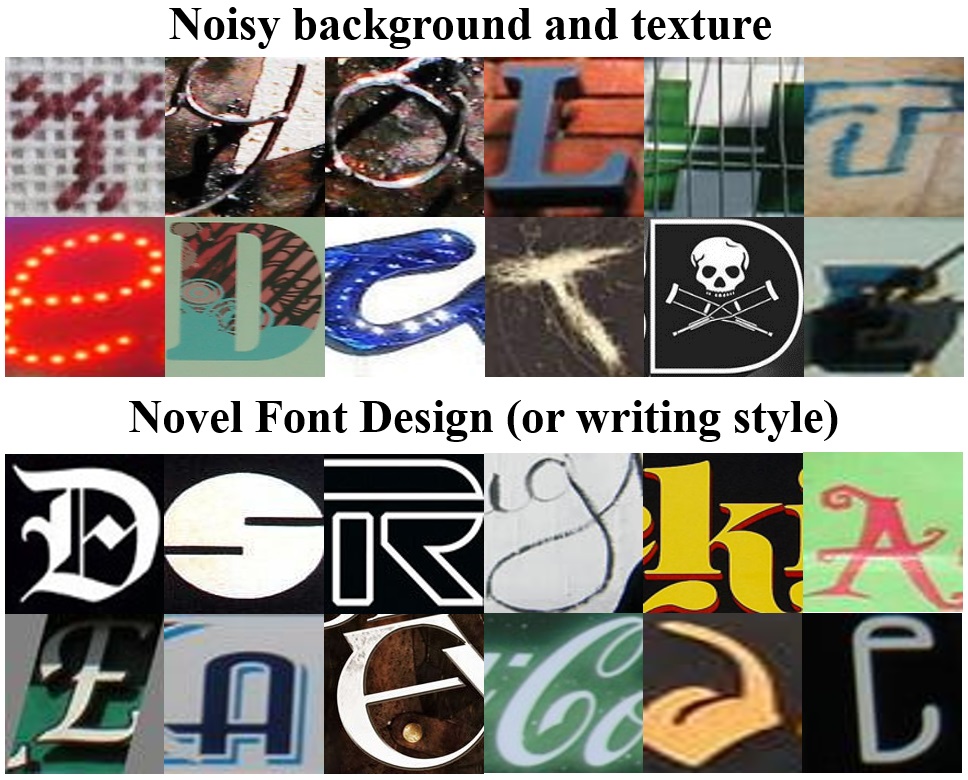}
  \caption{Scene character samples that existing CNN-based models typically fail to recognize.
  We classify the failure situations into two categories: one is noisy background and texture, the other is novel font design (or writing style).
  }
  \label{fig:CNNFailures}
\end{figure}

Through our experiments, we find that existing CNN-based models typically fail in handling the following two situations:
\begin{itemize}
\item \textbf{Noisy background and texture.}
CNNs employ convolutional filters to extract useful patterns and combine them for recognition.
However, noisy background and texture tend to distract CNNs from locating useful patterns accurately.
\item \textbf{Novel Font Design (or writing style).}
The human creativity continuingly endows characters with novel font designs (or writing styles).
As we know, even a small deformation of objects could lead off-the-shelf CNNs to failure in object recognition.
Similarly, existing CNN-based models perform poorly in recognizing those new images whose font styles differ a lot with training images.
Specific examples of these two situations are shown in Figure~\ref{fig:CNNFailures}.
\end{itemize}

\begin{figure}[t!]
  \centering
  \includegraphics[width=8cm]{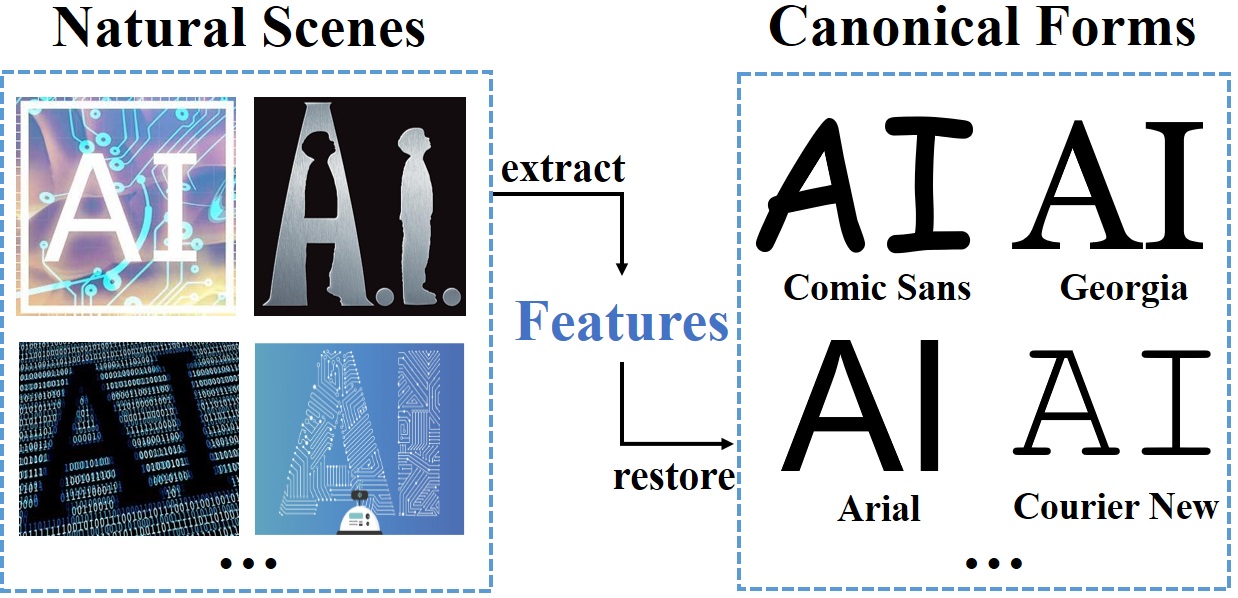}
  \caption{The main idea of our work.
  Extracted features of scene characters are endowed with the ability to portray their canonical forms of glyphs in multiple font styles.
  By adding this constraint, redundant information in features is reduced and thus recognition performance can be improved.}
  \label{fig:FontSamples}
\end{figure}

Existing methods ignore the following two important facts: First, glyphs in canonical forms can be adopted as helpful guidance for SCR in addition to character labels.
Most existing models employ character labels as the only guidance information.
In this paper, we define canonical forms of glyphs as the character images in some commonly-used and easy-to-read fonts, such as Arial, Times New Roman, Courier (English), Song and Kai (Chinese), etc.
Furthermore, it is easy to obtain these character images as they can be directly generated from corresponding font files.
Referring to canonical forms of glyphs is actually seeking for a mapping relation from scene characters to their standard glyphs.
This mapping relation can instruct models to further exclude useless scene information and concentrate on the shapes of characters.
Moreover, we believe that this process is accordant with the habit of human cognition and behavior.
\par

Second, font style information (letter endings, weights, slopes, etc.) is useless for recognizing character content and not supposed to be encoded into character features.
Existing models tend to fail in recognizing character images with new font styles, which are commonly-seen in SCR scenarios while few previous works have paid attention to that.
Accordingly, we define two important properties of features extracted from character images, which markedly affect the recognition performance:
\begin{itemize}
\item \textbf{Scene-independent.}
Features for character recognition are supposed to focus on the characters' shapes instead of scene information like background, illumination, special effects, etc.
In our method, the extracted deep features are forced to be capable of reconstructing standard printed glyphs, which is helpful to reduce useless scene information.
In this manner, our model acquires a better global perception of characters’ shapes so as to ignore noisy backgrounds and textures.
\item \textbf{Font-independent.}
Font information of characters ought not to be contained in extracted features for recognition.
It is necessary to decrease the font style information in extracted features to increase the robustness of models.
Our model is designed to generate glyphs of different font styles by decoupling content features and font style features.
Therefore, the learnt features are supposed to contain as less font information as possible during training stage.
When meeting characters with new font designs, our model could still find useful local patterns.

\end{itemize}
Motivated by the above-mentioned analyses, we propose a novel framework named Character Generation and Recognition Network (CGRN),
in which canonical forms of glyphs are used as prior knowledge for guiding models to learn scene-independent and font-independent features for SCR.
\section{Related Work}
\subsection{Image Synthesis}
Image synthesizing methods essentially fall into two categories: parametric and nonparametric.
The nonparametric models generate target images by copying patches from training images.
In recent years, parametric models based on Generative Adversarial Networks (GANs)~\cite{Goodfellow2014Generative} have been popular and achieved impressive results, such as LAPGAN~\cite{denton2015deep} and DCGAN~\cite{Radford2015Unsupervised}.
Image-to-image translation is a specific problem in the area of image synthesis.
Most recent approaches utilize CNNs to learn a parametric translation function by training a dataset of input-output examples.
Inspired by the success of GANs in generative tasks, the ``pix2pix'' framework~\cite{Isola2017Image} uses a conditional generative adversarial network to learn a mapping from input to output images.
Our key observation is that the learned mapping from a complex distribution (natural scene) to a simple distribution (canonical form) can help to handle the recognition task.
Motivated by this observation, generation and recognition functions are integrated together and work cooperatively in our model.
\subsection{Scene Character Recognition}
The problem of character recognition has attracted intensive attentions from AI/CV/PR communities for several decades. Traditional methods rely heavily on hand-crafted features (e.g., HOG~\cite{Dalal2005Histograms}) and typically fail to obtain satisfactory performance on scene character recognition (SCR). Recently, deep learning based approaches became as the predominant way to handle the SCR task. For instance, \cite{Wang2013End},~\cite{Bissacco2013PhotoOCR},~\cite{Jaderberg2014Deep} employed end-to-end CNNs to extract characters' features for recognition.
\cite{Wang2017Learning} proposed a model named SEDPD to improve the classification performance of CNN features by learning discriminative part detectors.
In handwritten character recognition (HCR), \cite{Zhang2018Robust} designed a adversarial feature learning (AFL) model to learn writer-independent features under the guidance of printed data. However, above mentioned methods are all classification-based feature learning frameworks while our model is based on generative models. Character labels are employed as their only guidance information in existing methods except AFL. Compared to AFL, our generative model aims to transfer the input scene character image to its canonical forms of glyphs, whose semantic and font information are both utilized by our method to excavate scene\&font-independent features.

\section{Method Description}
In this section, we describe the methodology of learning canonical forms of glyphs and present the details of our proposed network (i.e., CGRN).
\begin{figure*}[t!]
  \centering
  \includegraphics[width=17cm]{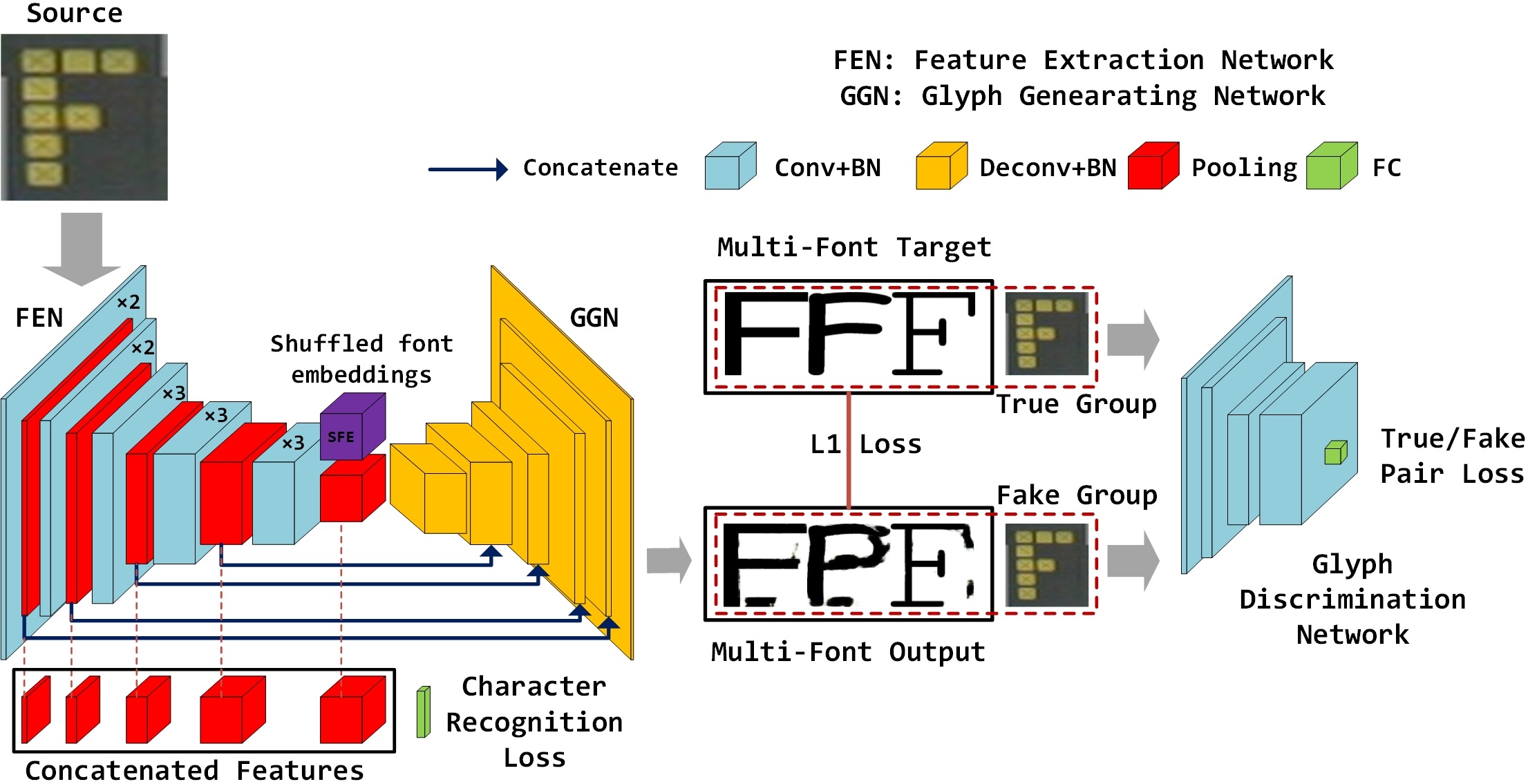}
  \caption{The architecture of our proposed network.}
  \label{fig:NetworkArch}
\end{figure*}
\subsection{Character Generation and Recognition Network}
CGRN builds a bridge between scene characters and standard glyphs to acquire a better understanding of their meanings.
Specifically, given an input character image $x$, CGRN tries to generate its corresponding glyphs of canonical forms in multiple font styles and predict its character label simultaneously.
Let us denote its character label as $y$ and its corresponding target images as $T = \{t_{1}, t_{2}, ..., t_{m}\}$,
where $y \in Y$, $Y = \{1, 2, ..., L\}$ is a finite label set, $L$ denotes the number of character classes, and $m$ is the number of font categories.
\subsection{Network Architecture}
As shown in Figure~\ref{fig:NetworkArch}, the proposed CGRN is composed of four subnetworks: Feature Extraction Network (FEN), Character Classification Network (CCN), Glyph Generating Network (GGN) and Glyph Discrimination Network (GDN). Details of the network architecture are presented in the following subsections.

\subsubsection{Feature Extraction Network}
The Feature Extraction Network is composed of a bunch of convolutional and pooling layers.
Given the input character image $x$, FEN learns a deep feature $E(x,\theta_{e})$, where $\theta_{e}$ denotes the parameters of FEN.
To describe characters' semantic information more precisely, multi-layer features are combined together to represent $E(x,\theta_{e})$:
\begin{equation}
E(x,\theta_{e}) = \{ E_{l_{1}}(x,\theta_{e}), E_{l_{2}}(x,\theta_{e}), ..., E_{l_{k}}(x,\theta_{e}) \},
\end{equation}
where $E_{l_{i}}(x,\theta_{e})$ denotes the output features  of $l_{i}$-th layer ($l_{1} < l_{2} < ... < l_{k}$).

\subsubsection{Glyph Generating Network}
The Glyph Generating Network takes the deep features of source images and font embeddings as input, and generates corresponding glyphs in canonical forms.
Specifically, the extracted deep features are utilized to generate character images in multiple font styles through a bunch of deconvolutional layers, as shown in the \emph{``Multi-Font Output''} part in Figure~\ref{fig:NetworkArch}. \par
Given the extracted features $E(x,\theta_{e})$ and a font embedding $z_{i}$, $x$'s corresponding canonical form $\hat{t}_{i}$ in font style ${f_{i}}$ can be generated by GGN
\begin{equation}
\hat{t}_{i} = G(E(x,\theta_{e}),z_{i}, \theta_{g}), 1 \leq i \leq m,
\end{equation}
where $z_{1}, z_{2}, ..., z_{m}$ are font category embedding vectors and $\theta_{g}$ denotes the parameters of GGN.
The generation of $\hat{t}_{1}, \hat{t}_{2}, ..., \hat{t}_{m}$ is parallel and the process is very similar to group convolution. \par
Inspired by the U-NET proposed by~\cite{Ronneberger2015U}, we add skip connections in the corresponding layers of FEN and GGN to reduce the information loss during the downsampling process (shown in Figure~\ref{fig:NetworkArch}).

\subsubsection{Character Classification Network}
The aim of our Character Classification Network (CCN) is to utilize the deep features extracted from input images to classify them into corresponding character categories.
First, multi-scale features in $E(x,\theta_{e})$ are down-sampled and concatenated as
\begin{equation}
E^{'}(x,\theta_{e}) = concat( E^{'}_{l_{1}}(x,\theta_{e}), ..., E^{'}_{l_{k}}(x,\theta_{e}) ),
\end{equation}
where $E^{'}_{l_{i}}(x,\theta_{e})$ is down-sampled from $E_{l_{i}}(x,\theta_{e})$, $1 \leq i \leq k$.
$E^{'}(x,\theta_{e})$ is then sent into CCN to compute its classification probability
\begin{equation}
P(y_{c} = y | x; \theta_{e}, \theta_{c}) = \frac{e^{-C_{y}({ E^{'}(x,\theta_{e}), \theta_{c}})}} {\sum_{j=1}^{L}{e^{-C_{j}({  E^{'}(x,\theta_{e}), \theta_{c}})}}},
\end{equation}
where $y_{c}$ denotes the predicted category label, $\theta_{c}$ denotes the parameters of CCN, and $C_{j}(\bullet)$ means the value of $j$-th element in the CCN's output vector.

\subsubsection{Glyph Discrimination Network}
The introduction of Glyph Discrimination Network (GDN) mainly borrows the idea of Generative Adversarial Networks (GANs) \cite{Goodfellow2014Generative}.
GAN has been proved to be able to markedly improve the quality of generated images.
In CGRN, GDN is employed to discriminate between generated (fake) glyphs and real glyphs.
Meanwhile, GGN is devoted to generate glyphs which are good enough to fool GDN.
Through this adversarial process, the quality of generated glyphs becomes better and better.
\par
GDN takes an image pair as input and predict its label $y_{d}$. The real pair $(x, t)$ corresponds to $y_{d} = 1$, while the fake pair $(x, \hat{t})$ corresponds to $y_{d} = 0$, where $t \in T = \{t_{1}, t_{2}, ..., t_{m}\}$, $\hat{t} \in \hat{T} = \{\hat{t}_{1}, \hat{t}_{2}, ..., \hat{t}_{m}\}$.
The probabilities of $y_{d} = 1$ and $y_{d} = 0$ predicted by GDN are defined as

\begin{equation}
{P(y_{d} = 1 | x, t_{i}; \theta_{d})} = \frac{1}{1+e^{-D(x, t_{i}, \theta_{d})}},
\end{equation}

\begin{equation}
P(y_{d} = 0 | x, t_{i}; \theta_{d}) =  1 - P(y_{d} = 1 | x, t_{i}; \theta_{d}),
\end{equation}

\begin{equation}
{P(y_{d} = 1 | x, \hat{t_{i}}; \theta_{d})} = \frac{1}{1+e^{-D(x, \hat{t_{i}}, \theta_{d})}},
\end{equation}

\begin{equation}
P(y_{d} = 0 | x, \hat{t}_{i}; \theta_{d}) =  1 - P(y_{d} = 1 | x, \hat{t_{i}}; \theta_{d}),
\end{equation}
where $\theta_{d}$ denotes the parameters of GDN, $1 \leq i \leq m$ and $D(\bullet)$ means the value of GDN's output.
\subsubsection{Detailed Configurations}
The detailed configurations of CGRN are shown in Table~\ref{tb:NetworkParams}. The first, second and third columns show network layers' names, types and parameters, respectively.
Thereinto, ``Conv'' means convolutional layer, ``Pool'' means pooling layer, ``Deconv'' means deconvolutional layer and ``FC'' means fully-connected layer.
For convolutional and deconvolutional layers, ``$k \times k \times c $, $s$, BN, ReLU'' in each row denotes that the kernel size is $k$, the stride is $s$, the number of output features' channels is $c$, activation function ReLU and batch normalization~\cite{Ioffe2015Batch} are employed.
For pooling layers, ``$k \times k $, $s$'' in each row denotes that the pooling kernel size is $k$, the stride is $s$.
For fully-connected layers, ``$i \times o$'' in each row describes the input dimension and output dimension of features.
The configuration of FEN is basically consistent with convolutional layers of VGG16~\cite{Simonyan2014Very}, except the last pooling layer and batch normalization for each convolutional layer.
Note that the architectures of our FEN and GGN are asymmetrical, which is different against the original UNET, as we expect FEN goes deeper to acquire richer features of scene characters.
\begin{table}[!htbp]
\centering
\caption{Detailed Configurations of our CGRN.}
\label{tb:NetworkParams}
\scalebox{0.8}{
\begin{tabular}{|c|c|c|}
\hline
\multicolumn{3}{|c|}{Input: $ 64  \times 64$ RGB image}\\
\hline
Layer & Type &Parameters\\
\hline
\multicolumn{3}{|c|}{Feature Extraction Network}\\
\hline
$E\_conv1$ ($\times 2$) & Conv & $3 \times 3  \times 64$, $1$, BN, ReLU \\
\hline
$E\_pool1$              & Pooling & $2 \times 2$, $2$ \\
\hline
$E\_conv2$ ($\times 2$) & Conv & $3  \times 3 \times 128$, $1$, BN, ReLU \\
\hline
$E\_pool2$              & Pooling & $2 \times 2$, $2$ \\
\hline
$E\_conv3$ ($\times 3$) & Conv  & $3  \times 3 \times 256$, $1$,BN, ReLU \\
\hline
$E\_pool3$   & Pooling & $2 \times 2$, $2$ \\
\hline
$E\_conv4$ ($\times 3$) & Conv & $3 \times 3 \times 512$, $1$,BN, ReLU \\
\hline
$E\_pool4$              & Pooling & $2 \times 2$, $2$ \\
\hline
$E\_conv5$ ($\times 3$) & Conv & $3 \times 3 \times 512$, $1$,BN, ReLU \\
\hline
$E\_pool5$  & Pooling &  $4 \times 4$, $4$ \\
\hline
\multicolumn{3}{|c|}{Character Classification Network}\\
\hline
$C\_pool1$              & Pooling & $32 \times 32$, $32$ \\
\hline
$C\_pool2$              & Pooling & $16 \times 16$, $16$ \\
\hline
$C\_pool3$              & Pooling & $8 \times 8$, $8$ \\
\hline
$C\_pool4$              & Pooling & $4 \times 4$, $4$ \\
\hline
$C\_pool5$              & Pooling & $1 \times 1$, $1$ \\
\hline
$C\_fc$  & FC &  $1472 \times L $   \\
\hline
\multicolumn{3}{|c|}{Glyph Generating Network}\\
\hline
$G\_deconv1,2$ & Deconv & $5 \times 5 \times 512$, $2$, BN, ReLU \\
\hline
$G\_deconv3$ & Deconv & $5 \times 5 \times 256$, $2$, BN, ReLU \\
\hline
$G\_deconv4$  & Deconv & $5 \times 5 \times 128$, $2$, BN, ReLU \\
\hline
$G\_deconv5$ & Deconv & $5 \times 5 \times 64$, $2$, BN, ReLU \\
\hline
$G\_deconv6$  & Deconv & $5 \times 5 \times 3$, $2$, BN, ReLU \\
\hline
\multicolumn{3}{|c|}{Glyph Discrimination Network}\\
\hline
$D\_conv1$  & Conv & $5 \times 5 \times 64$, $2$, BN, ReLU \\
\hline
$D\_conv2$  & Conv & $5 \times 5 \times 128$, $2$, BN, ReLU \\
\hline
$D\_conv3$  & Conv & $5 \times 5 \times 256$, $2$, BN, ReLU \\
\hline
$D\_conv4$  & Conv & $5 \times 5 \times 512$, $1$, BN, ReLU \\
\hline
$D\_fc$  & FC & $32,768 \times 1$\\
\hline
\end{tabular}
}
\end{table}
\subsection{Learning Scene\&Font-Independent Features}
Although CNN based models have already achieved impressive performance in removing useless scene information for character recognition, we believe their performance can be further improved by adding the guidance of glyphs in canonical forms.
As shown in Figure~\ref{fig:NetworkArch}, the target images are composed of clear backgrounds and corresponding glyphs in canonical forms.
The aim of the proposed CGRN is to learn scene\&font-independent features by reconstructing these character images accurately.\par

As we know, glyphs in any kinds of font styles might be contained in a given natural scene image. If we mix semantic features with font information, our model might fail to recognize those characters whose font styles have never been seen in training images.
Therefore, the extracted character features $E(x,\theta_{e})$ are supposed to be font-independent in addition to scene-independent.
By introducing the font embedding mechanism into CGRN, we expect to separate font features from semantic features, with $z$ representing the former and $E(x,\theta_{e})$ for the latter.
Furthermore, by setting multi-font glyphs as target, the extracted features are forced to contain as less font information as possible, so as to reconstruct glyphs in various font styles.
However, if we set single-font glyphs as target, font information is hardly to be removed especially when many training characters' fonts share similar style with the target font.
In our work, we select English and Chinese characters in representative fonts as the target glyphs in canonical forms for CGRN to generate, details are presented in the following sections.
\par

\subsection{Loss Function}
In CGRN, we define three loss functions.
The first one (i.e., the pixel loss $L_{pixel}$), which measures the dissimilarity between generated character images $\hat{t}$ and their corresponding glyphs of canonical forms ${t}$ all over ${m}$ fonts, is defined as
\begin{equation}
\begin{split}
L_{pixel} & =  \mathbb{E}_{t, \hat{t}}[\frac{1}{m}{\sum_{i=1}^{m}{\Vert\hat{t}_{i} - t_{i}\Vert}}] \\
& =  \mathbb{E}_{x, t, z}[\frac{1}{m}{\sum_{i=1}^{m}{\Vert G(E(x,\theta_{e}),z_{i}, \theta_{g}) - t_{i}\Vert}}].
\end{split}
\end{equation}
The character recognition loss $L_{CR}$, which indicates the recognition error of CCN, is defined as
\begin{equation}
L_{CR} = -\mathbb{E}_{x, y}[\log(P(y_{c} = y | x; \theta_{e}, \theta_{c}))].
\end{equation}
The discriminator loss $L_{D}$, which indicates the identification error of GDN all over ${m}$ image pairs, is defined as
\begin{equation}
\begin{split}
L_{D} =
-\mathbb{E}_{x, t}[\frac{1}{m}\sum_{i=1}^{m}\log(P(y_{d} = 1 | x, t_{i}; \theta_{d}))] \\
- \mathbb{E}_{x, \hat{t}}[\frac{1}{m}\sum_{i=1}^{m}\log(P(y_{d} = 0 | x, \hat{t}_{i}; \theta_{d}))].
\end{split}
\end{equation}

\subsection{Training Process}
We optimize the whole network by introducing GDN to play the minmax game with FEN, GGN, and CCN:
\begin{equation}
\min \limits_{\theta_{e},\theta_{c}, \theta_{g}} \max \limits_{\theta_{d}} \lambda L_{CR}(\theta_{e},\theta_{c}) +  \lambda L_{pixel}(\theta_{e}, \theta_{g}) - L_{D}(\theta_{e}, \theta_{g}, \theta_{d}),
\end{equation}
where $\lambda$ is a fixed weight coefficient (set as 100 in our experiments).
Our optimization strategy for this objective function is training FEN, CCN, GGN and GDN alternatively, which is similar to~\cite{Goodfellow2014Generative}. Details are as follows.\par
In each mini-batch, we first optimize GDN:
\begin{equation}
\theta_{d} \leftarrow \theta_{d} - \mu\frac{\partial L_{D}}{\partial\theta_{d}}.
\label{equ:thetad}
\end{equation}
Then, we optimize FEN, GGN and CCN jointly:
\begin{equation}
\theta_{e}^{'} \leftarrow \theta_{e} - \mu(\frac{\partial \lambda L_{CR}}{\partial\theta_{e}} + \frac{\partial \lambda L_{pixel}}{\partial\theta_{e}} - \frac{ \partial L_{D}}{\partial\theta_{e}}),
\label{equ:thetae}
\end{equation}
\begin{equation}
\theta_{g}^{'} \leftarrow \theta_{g} - \mu(  \frac{\partial \lambda L_{pixel}}{\partial\theta_{g}} - \frac{ \partial L_{D}}{\partial\theta_{g}}),
\label{equ:thetag}
\end{equation}
\begin{equation}
\theta_{c}^{'} \leftarrow \theta_{c} - \mu\frac{\partial \lambda L_{CR}}{\partial\theta_{c}},
\label{equ:thetac}
\end{equation}
\begin{equation}
\theta_{e} \leftarrow \theta_{e}^{'} , \theta_{g}  \leftarrow \theta_{g}^{'},  \theta_{c}  \leftarrow \theta_{c}^{'},
\end{equation}
where $\mu$ is the learning rate used for training model.
Note that Equation~\ref{equ:thetad},~\ref{equ:thetae},~\ref{equ:thetag} and~\ref{equ:thetac} are presented in the basic form of gradient descent for brevity.
Practically, we employ a gradient-based optimizer with adaptive moment estimation (i.e., Adam optimizer~\cite{Kingma2014Adam}).
\section{Experiments}
To verify the effectiveness and generality of our model, we conduct experiments on datasets of three widely-used languages: English, Chinese and Bengali, which represent alphabet, logography and abugida, respectively.
We also design comparative experiments to demonstrate the effect of proposed techniques in our model.
\subsection{Datasets}

English character datasets include \textbf{ICDAR 2003}~\cite{Lucas2003ICDAR} and \textbf{IIIT5K}~\cite{Mishra2013Scene} dataset.
They all contain English letters in 52 classes and Arabic numbers in 10 classes (i.e., a-z, A-Z, 0-9).
\begin{itemize}
\item The task on the ICDAR 2003 character dataset is very challenging because of serious non-text background outliers with cropped character samples, and many character images have very low resolution.
      It contains 6,113 character images for training and 5,379 for testing
\item The IIIT5K dataset consists of 9,678 character samples for training and 15,269 for testing. The dataset contains both scene text images and born-digital images.
\end{itemize}
Chinese character dataset : \textbf{Pan+ChiPhoto} dataset~\cite{Tian2016Multilingual}.
It is built by the combination of two datasets: \textbf{ChiPhoto} and \textbf{Pan\_Chinese\_Character} dataset.
The images in this dataset are mainly captured at outdoors in Beijing and Shanghai, China, which involve various scenes like signs, boards, advertisements, banners, objects with texts printed on their surfaces.
In~\cite{Tian2016Multilingual}, ChiPhoto was split into two parts (60\% and 40\%) and added to the training and testing datasets of Pan\_Chinese\_Character.
Because the authors did not give the exact split of training and testing images, we follow their steps to split these images again.
Finally, we have 6098 training images and 4220 testing images, totally 1203 character classes. \par
Bengali character dataset: \textbf{ISI\_Bengali\_Character} dataset \cite{Tian2016Multilingual}.
Bengali script is the 6th most popularly used script in the world and it holds official status in the two neighboring countries Bangladesh and India.
This dataset contains 158 classes of Bengali numerals, characters or their parts.
19,530 Bengali character samples are divided into two sets in which 13,410 images are used for training and 6,120 for testing. \par
\begin{figure}[t!]
  \centering
  \includegraphics[width=8cm]{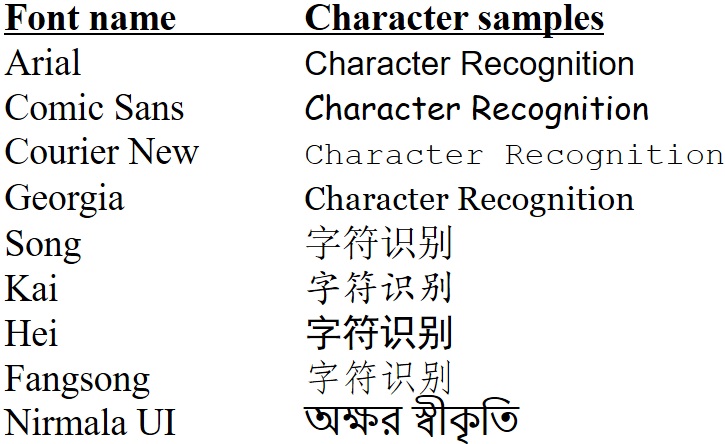}
  \caption{Samples of characters rendered by our selected fonts.}
  \label{fig:FontCharactersSamples}
\end{figure}

Selected fonts for target canonical forms of glyphs: In our experiments, we select 4 fonts for English glyphs: \textbf{Arial}, \textbf{Comic Sans}, \textbf{Courier New} and \textbf{Georgia}, 4 fonts for Chinese glyphs: \textbf{Song}, \textbf{Kai}, \textbf{Hei} and \textbf{Fangsong}, and one font for Bengali glyphs: \textbf{Nirmala UI}.
Character samples rendered by these fonts are shown in Fig~\ref{fig:FontCharactersSamples}.
For English and Chinese, we select these fonts based on their popularity and style diversity.
We only select one font for Bengali glyphs because we are not familiar with the language.
\subsection{Data Preprocessing}
All character images (including source images and target images) are resized into the resolution of $64\times64$.
Besides shuffling training data images, font embeddings are also shuffled for each training image $x$ to avoid overfitting:
 $(z_{1}, z_{2}, ..., z_{m}) \rightarrow (z_{p_{1}}, z_{p_{2}}, ..., z_{p_{m}})$, where $p_{1}, p_{2}, ..., p_{m}$ is one permutation of $1, 2, ..., m$.
Accordingly, the target images $(t_{1}, t_{2}, ..., t_{m})$ are shuffled to $(t_{p_{1}}, t_{p_{2}}, ..., t_{p_{m}})$.
\subsection{Implementation Details}
The parameters of FEN are initialized by the pre-trained VGG16~\cite{Simonyan2014Very} on ImageNet dataset~\cite{Deng2009ImageNet}, and the others are initialized by using random weights with normal distribution of $0$ mean and $0.02$ standard deviation.
We adopt Adam optimizer~\cite{Kingma2014Adam} to train our model where the initial learning rate $\mu$ is set to $0.0001$ and the first order of momentum is $0.5$.
The batch size of training images is set to $128$.
Our method is implemented under the TensorFlow framework~\cite{GTF}.

\begin{table}[!htbp]
\centering
\caption{Character recognition results of different methods evaluated on the English Datasets}
\label{tb:ResOfEngDatasets62Classes}
\begin{tabular}{|p{3.5cm}|p{1cm}|p{1cm}|}
\hline
  Method                                & IC03                          & IIIT5K   \\
\hline
  HOG + SVM~\cite{Dalal2005Histograms}   & 77.0                          & 70.0      \\
\hline
  CNN~\cite{Jaderberg2014Deep}          & 86.8                          & 78.5      \\
\hline
  RPCA-SR~\cite{zhang2016natural}     & 79.0                          & 76.0      \\
\hline
  CO-HOG~\cite{Tian2016Multilingual}  & 80.5                           & 77.8      \\
\hline
  ConvCoHOG~\cite{Tian2016Multilingual}  & 81.7                           & 78.8      \\
\hline
  SEDPD~\cite{Wang2017Learning}      & 84.0                        & 80.3     \\
\hline
  VGG16 (pretrained)              & 84.5                        & 82.1     \\
\hline
  CGRN (-GGN, -GDN)                  & 85.5                        & 84.9 \\
\hline
  CGRN (SF, -GDN)                &  86.3                  & 85.2      \\
\hline
  CGRN (MF, -GDN)                &  86.9                  &       85.5 \\
\hline
  CGRN (SF)                   &  86.8                               &   85.5     \\
\hline
  CGRN (MF)                  &   \textbf{87.1}              &  \textbf{85.6}     \\
\hline
\end{tabular}
\end{table}

\begin{table}[!htbp]
\centering
\caption{Character recognition results of different methods evaluated on the Chinese Dataset}
\label{tb:ResOfChnDatasets}
\begin{tabular}{|p{4cm}|p{2cm}|}
\hline
  Method                       & Pan+ChiPhoto  \\
\hline
  Tesseract OCR~\cite{GTOCR}                & 20.5               \\
\hline
  HOG                          &  59.2             \\
\hline
  CNN~\cite{Jaderberg2014Deep}                          &  61.5            \\
\hline
  CNN +SVM~\cite{Jaderberg2014Deep}                     & 62.3              \\
\hline
   ConvCoHOG\_NewOffset~\cite{Tian2016Multilingual}  &   71.2             \\
\hline
  VGG16 (pretrained)                    &   85.8           \\
\hline
  CGRN (-GGN, -GDN)                  & 87.5                     \\
\hline
   CGRN (SF, -GDN)            &  88.6             \\
\hline
  CGRN (MF, -GDN)            &   88.9           \\
\hline
  CGRN (SF)            &   89.1             \\
\hline
  CGRN (MF)           &    \textbf{89.4}           \\
\hline
\end{tabular}
\end{table}

\begin{table}[!htbp]
\centering
\caption{Character recognition results of different methods evaluated on the Bengali Dataset}
\label{tb:ResOfBegDatasets}
\begin{tabular}{|p{4cm}|p{2cm}|}
\hline
  Method                           & ISI\_Bengali   \\
\hline
  Tesseract OCR~\cite{GTOCR}       & -                       \\
\hline
  HOG                              & 87.4                      \\
\hline
  CNN~\cite{Jaderberg2014Deep}      & 89.7                      \\
\hline
  CNN+SVM~\cite{Jaderberg2014Deep}   & 90.1                      \\
\hline
 ConvCoHOG\_NewOffset~\cite{Tian2016Multilingual}         & 92.2                \\
\hline
  VGG16 (pretrained)              & 96.3                  \\
\hline
  CGRN (-GGN, -GDN)                  & 96.9        \\
\hline
  CGRN (-GDN)                        & 97.1       \\
\hline
  CGRN                          &\textbf{97.4}        \\
\hline
\end{tabular}
\end{table}
\subsection{Comparison with Other Methods}
Table~\ref{tb:ResOfEngDatasets62Classes},~\ref{tb:ResOfChnDatasets}, and~\ref{tb:ResOfBegDatasets} show different methods' performance on aforementioned datasets.
As the parameters of FEN are initialized by the pre-trained VGG16, we add a fine-tuned VGG16 model into comparison for the sake of fairness.
In these tables, ``SF'' denotes the best performing version of CGRN when employing single-font glyphs of canonical forms as target.
``-GDN'' denotes the CGRN model in which GDN is removed.
``-GGN'' denotes the CGRN model in which GGN is removed.
``MF'' means employing multi-font glyphs as target in CGRN (we experimentally set $t=4$ in English and Chinese datasets).
As shown in these tables, our method clearly outperforms other existing approaches. \par
For the experiments conducted on IC03 (ICDAR 2003), ~\cite{Jaderberg2014Deep} and ~\cite{Tian2016Multilingual} both used additional character images (107k and 18k) to train their models. On the contrary, our model uses the least training character images (6k) and achieves the best performance (see Table~\ref{tb:ResOfEngDatasets62Classes}).
Although utilizing the parameters of pre-trained VGG16, our model still possesses the advantages of high efficiency and better extensibility.
As is known to all, recognizing Chinese characters is a very challenging task on account of their large number of classes and quite complicated geometric structures.
Existing methods perform poorly on the Pan+ChiPhoto dataset.
VGG16 network greatly improves the recognition accuracy, owing to the pre-training on the ImageNet dataset.
By learning canonical forms of glyphs, our model is able to extract more discriminative features for Chinese characters in natural scene images and thus significantly improves the recognition accuracy (see Table~\ref{tb:ResOfChnDatasets}).
ISI\_Bengali\_Character, which contains many synthesized images, is a less challenging dataset compared to the others.
Our model mainly contributes to improving the recognition accuracy of scene characters in this dataset, so the improvement of the overall recognition accuracy is not that significant (see Table~\ref{tb:ResOfBegDatasets}).

\subsection{The Improvement of Image Feature Learning}

\begin{figure}[t!]
  \centering
  \includegraphics[width=8.3cm]{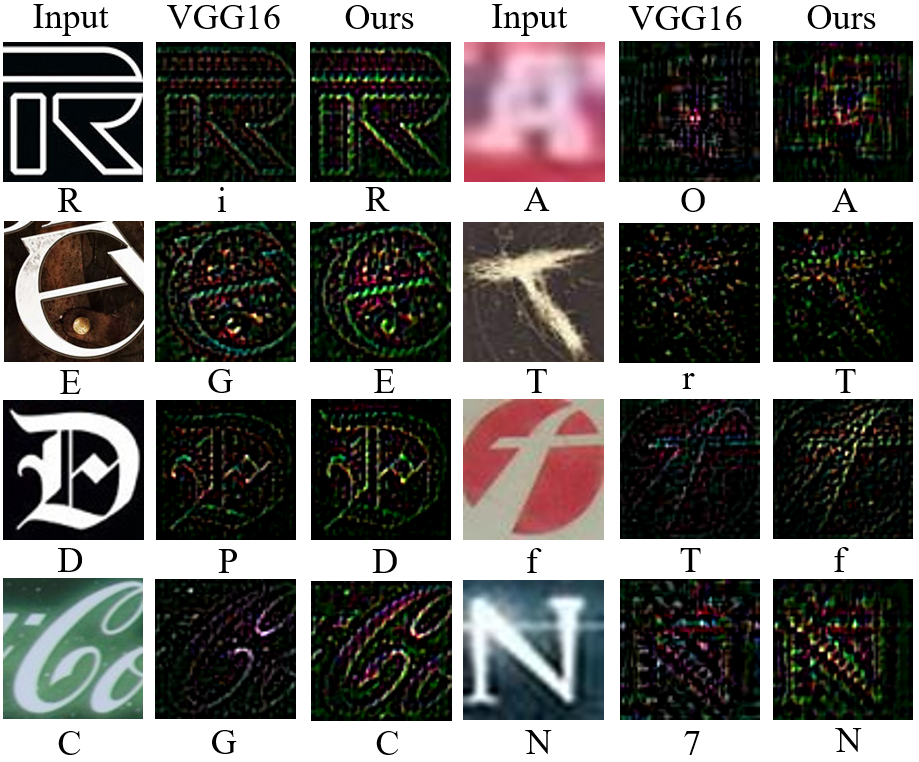}
  \caption{Visualization of CNN feature maps of the 5th pooling layer from VGG16 and our model.
  The light pixels reveal the structures of input image that excite CNN feature maps.
  The characters below each set of images are ground-truth label, VGG16's predicted label and our model's predicted label, respectively.
  Our model captures more useful local patterns of scene characters compared to VGG16.}
  \label{fig:VisOfFeat}
\end{figure}

In this section we give an illustration of how the feature learning is improved by learning canonical forms of glyphs.
We select VGG16 network for comparison, which is commonly used as feature extractor in many existing models~\cite{jaderberg2016reading,shi2017an,lee2016recursive}.
Utilizing the technique proposed by~\cite{zeiler2014visualizing}, we project the features activations back to the input pixel space, which is presented in Figure~\ref{fig:VisOfFeat}.
The light regions in the reconstructed images are responsible for the activations in extracted feature maps.
The selected cases in Figure~\ref{fig:VisOfFeat} are representative of the two situations mentioned in Section 1.
The left four cases possess novel font designs and the right four cases are placed in noisy backgrounds or have noisy textures.
Compared to VGG16, our model shows more accurate perception on the discriminative patterns of scene characters and correctly recognize them.
Our model is trained to capture the most discriminative local patterns so as to reconstruct multi-font canonical glyphs, which is the key to success.
Taking the letter ``E'' for example, our model pays more attention to the middle horizontal line of ``E'', which avoids recognizing it as ``G''.
Another example is the letter ``f'', our model's concentration on the head part of ``f'' leads to the successful recognition.
Learning to generate canonical forms of glyphs instructs our model to exclude undesired nuisance factors and concentrate on those useful patterns.

\subsection{Contribution of Glyphs Discrimination Network}
\begin{figure}[t!]
  \centering
  \includegraphics[width=8cm]{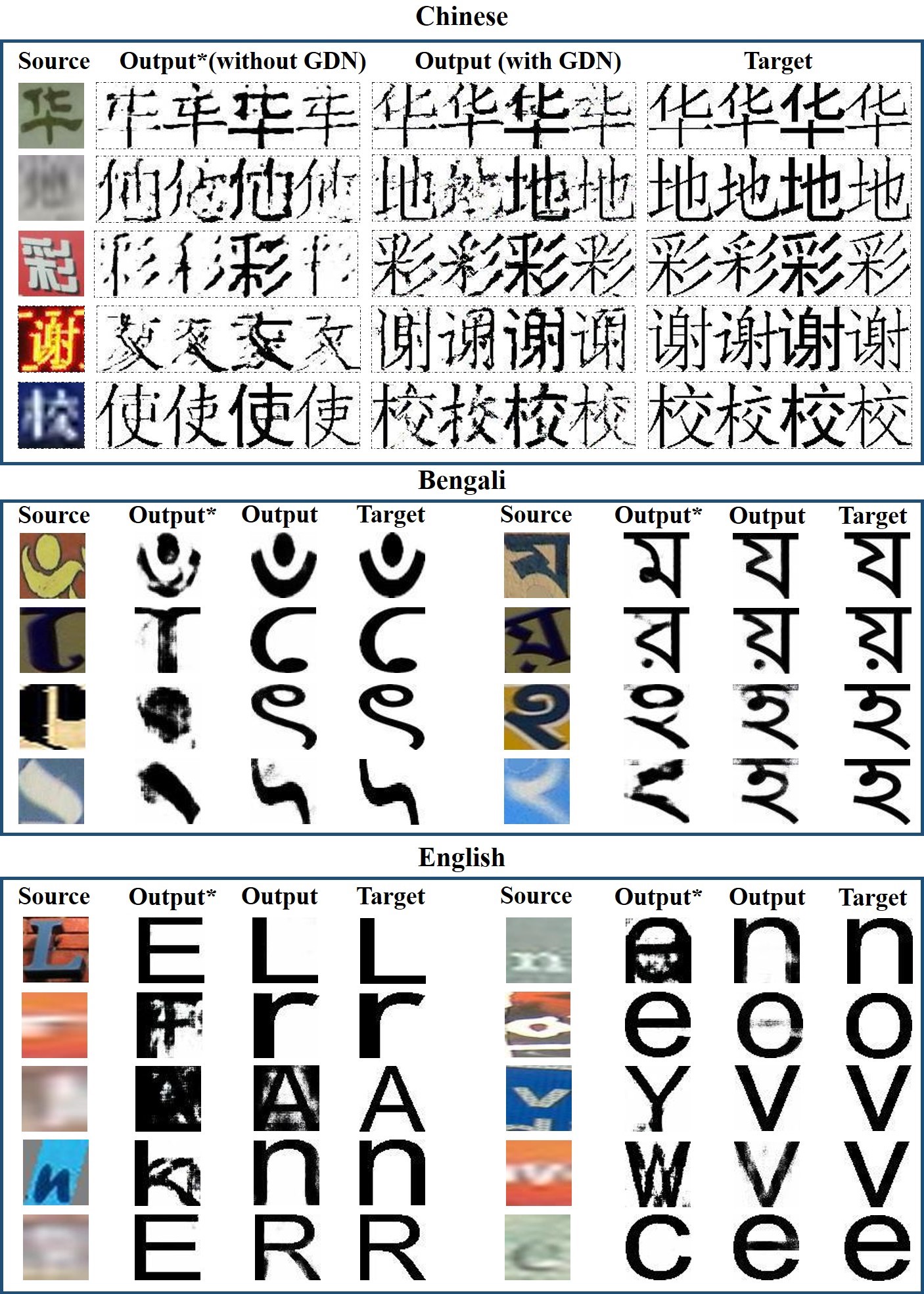}
  \caption{Samples of testing images which were wrongly classified without GDN but correctly classified after introducing GDN. Glyphs in the \emph{``output*''} and \emph{``output''} column are generated by CGRN (no GDN) and CGRN respectively.}
  \label{fig:EffectOfGAN}
\end{figure}

Glyphs Discrimination Network is proved to be effective in improving the quality of generated glyphs through our experiments (see Figure~\ref{fig:EffectOfGAN}).
Without the supervision of GDN, FEN and GGN tend to generate incorrect or unrecognizable glyphs when meeting blurry, distorted and ambiguous character images.
After introducing GDN, the generated images become more accordant with target images and easier to be recognized, although there still exist some flaws on them.
The improvement demonstrates that features learned by model with GDN are able to more precisely capture semantics and eliminate other interference information.
Consequently, the recognition accuracy is also improved (shown in Table~\ref{tb:ResOfEngDatasets62Classes},~\ref{tb:ResOfChnDatasets}, and~\ref{tb:ResOfBegDatasets}).
Character images in Figure~\ref{fig:EffectOfGAN} were wrongly classified without GDN but correctly classified after introducing GDN.
Through improving the quality of generated images, the learned deep features are equipped with enhanced expressiveness and become more scene-independent. \par
\subsection{Contribution of Multiple-Font Guidance}
\begin{figure}[t!]
  \centering
  \includegraphics[width=8cm]{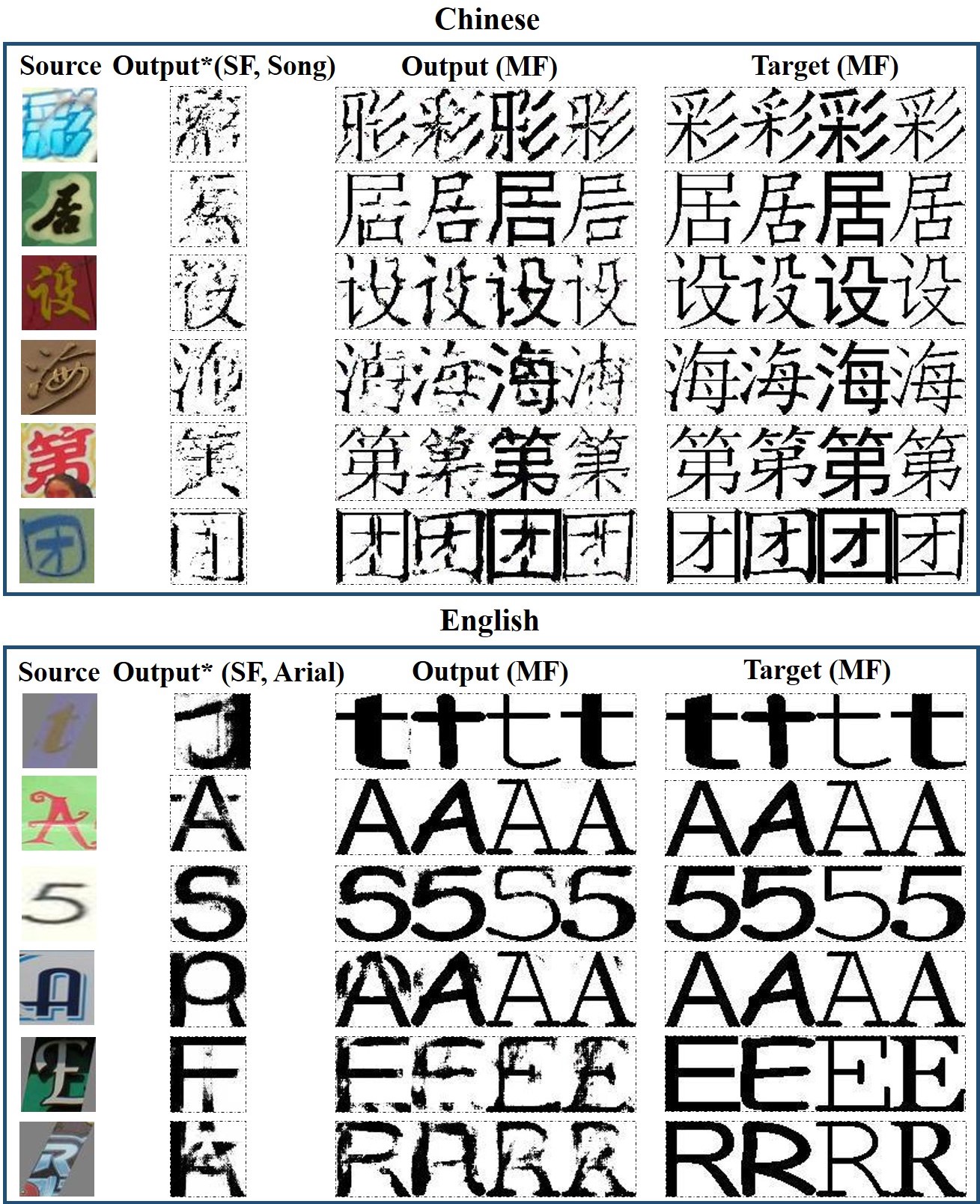}
  \caption{Samples of testing images which were wrongly classified without MFG but correctly classified after introducing MFG. Glyphs in the \emph{``output*''} and \emph{``output''} column are generated by CGRN (SF, GDN) and CGRN (MF, GDN) respectively.}
  \label{fig:EffectOfMFG}
\end{figure}

Our model becomes less sensitive to font variance under the multiple-font guidance (MFG) compared to the single-font guidance (SFG).
When we set multi-font glyphs as generation targets, there must be at least one glyph whose font style is different from the input scene character. To preciously reconstruct them all, the feature extractor must learn to capture the most discriminative local patterns for every training example.
By contrast, the training process of single-font model is not so sufficient as multi-font model, which leads single-font model more sensitive to the font variance.
When we adopt a single font as the target, our model tends to generate incorrect glyphs if the font style of input character images differs a lot with the target font style, as is shown in Figure~\ref{fig:EffectOfMFG}.
It is caused by the poor robustness of extracted features mixed with a lot of unnecessary font information.
But the character images in Figure~\ref{fig:EffectOfMFG} can be correctly classified when training our model with multiple-font guidance.
Multiple-font guidance instructs our model to distinguish semantics from font styles.
Table~\ref{tb:ResOfEngDatasets62Classes},~\ref{tb:ResOfChnDatasets}, ~\ref{tb:ResOfBegDatasets} and Figure~\ref{fig:EffectOfMFG} show that MFG further enhances our model's recognition and generation performance.
Our model understands characters more deeply by reconstructing them into canonical forms of glyphs in multiple fonts according to font embeddings.

\subsection{Experiments on Text String Recognition}
\begin{figure*}[t!]
  \centering
  \includegraphics[width=17cm]{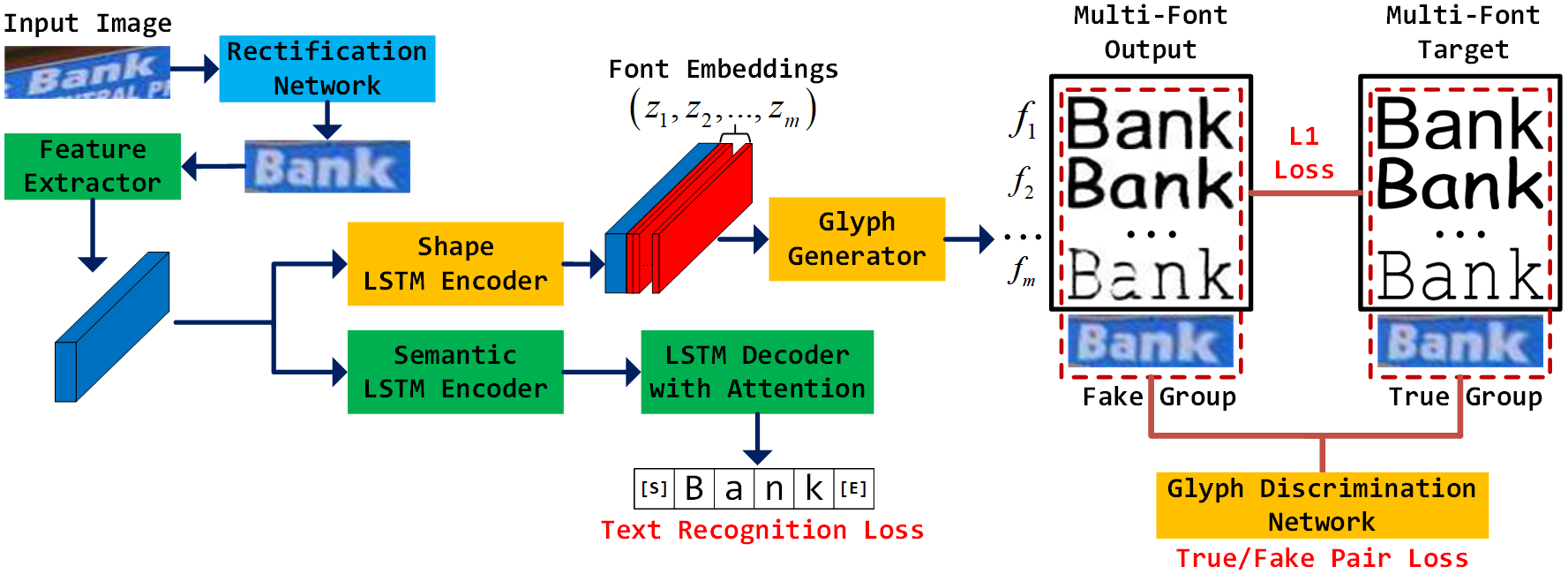}
  \caption{The architecture of our proposed network (ASTER+CGN) for text string recognition.}
  \label{fig:NetworkArchTSR}
\end{figure*}

The latest scene text recognition methods~\cite{shi2017an,cheng2017focusing,cheng2018aon,shi2018aster} are mainly designed to recognize text strings without explicitly splitting characters by combining Recurrent Neural Networks (RNNs) with CNNs.
To further verify our method’s effectiveness, we conduct experiments on text string recognition in addition to single character recognition.
Specifically, we add the proposed glyph generating and discrimination network to ASTER proposed by Shi \emph{et al.}~\cite{shi2018aster}.
The detailed architecture of this new model, named as ASTER+CGN (Character Generation Network), is presented in Figure~\ref{fig:NetworkArchTSR}.
ASTER consists of a text rectification network and a text recognition network, corresponding to the blue and green modules in Figure~\ref{fig:NetworkArchTSR}, respectively.
The newly added modules by us (marked as orange) consist of a LSTM encoder (for sequence modeling), a glyph generating network (i.e., glyph generator in Figure~\ref{fig:NetworkArchTSR}) and a glyph discrimination network.
We employ the extracted CNN features to generate canonical forms of a horizontal text, following the main idea of CGRN.
We add another dataset ICDAR-2013~\cite{karatzas2013icdar} into the experiment.
As shown in Table~\ref{tb:ResOfTSEngDatasets}, our method significantly improves the recognition accuracy of ASTER and achieves state-of-the-art performance.
Considering that the benefits of CGN have been fully discussed in previous sections, we will not show more experimental results here.

\begin{table}[!htbp]
\centering
\caption{Text string recognition results of different methods evaluated on benchmarks in lexicon-free mode.}
\label{tb:ResOfTSEngDatasets}
\begin{tabular}{|p{3cm}|p{1cm}|p{1cm}|p{1cm}|}
\hline
  Method                                & IC03          & IIIT5K    & IC13 \\
\hline
  Jaderberg \emph{et al.}~\cite{jaderberg2015deep}  & 89.6    & -    & 81.8 \\
\hline
  Jaderberg \emph{et al.}~\cite{jaderberg2016reading}  & 93.1    & -    & 90.8 \\
\hline
  Lee \emph{et al.}~\cite{lee2016recursive}  & 88.7    & 78.4    & 90.0 \\
\hline
  Shi \emph{et al.}~\cite{shi2017an}         & 91.9          & 81.2   & 89.6 \\
\hline
  Cheng \emph{et al.}~\cite{cheng2017focusing}   & 94.2         & 87.4    & 93.3  \\
\hline
  Cheng \emph{et al.}~\cite{cheng2018aon}   & 91.5         & 87.0    & -  \\
\hline
  Shi \emph{et al.}~\cite{shi2018aster}        & 94.5          & 93.4   & 91.8 \\
\hline
  ASTER+CGN                             &  \textbf{95.3}       &  \textbf{94.0}  & \textbf{94.4}  \\
\hline
\end{tabular}
\end{table}
\section{Towards Common Object Recognition}
In this paper, our work focuses on scene character recognition in natural scenes.
It is worth mentioning that the main idea of our work is not only limited to text recognition but also applicable to common object recognition tasks, such as face recognition with different poses, speech recognition with accents and so on.
Canonical forms of objects, which contribute to eliminate the interference factors in scene images, are more meaningful than any other object properties summarized by human.
Therefore, we believe that canonical forms of objects could be used as more complete labels than traditional numeric labels in many applications of object recognition.

\section{Conclusion}
This paper presented a deep learning based framework to boost the performance of SCR by learning canonical forms of glyphs.
The main contribution of the paper is to utilize canonical forms of glyphs as guidance to excavate scene-independent and font-independent features.
Our model also benefits from the adversarial learning process introduced in GAN.
The effectiveness and generality of our techniques were verified through experiments conducted on multilingual datasets.
Last but not the least, it is easy to extend our approach with a few changes to improve the performance of methods in many other object recognition tasks.

\bibliographystyle{spmpsci}
\bibliography{myreference}

\begin{thebibliography}{10}
\providecommand{\url}[1]{{#1}}
\providecommand{\urlprefix}{URL }
\expandafter\ifx\csname urlstyle\endcsname\relax
  \providecommand{\doi}[1]{DOI~\discretionary{}{}{}#1}\else
  \providecommand{\doi}{DOI~\discretionary{}{}{}\begingroup
  \urlstyle{rm}\Url}\fi

\bibitem{Bissacco2013PhotoOCR}
Bissacco, A., Cummins, M., Netzer, Y., Neven, H.: Photoocr: Reading text in
  uncontrolled conditions.
\newblock In: IEEE International Conference on Computer Vision, pp. 785--792
  (2013)

\bibitem{cheng2017focusing}
{Cheng}, Z., {Bai}, F., {Xu}, Y., {Zheng}, G., {Pu}, S., {Zhou}, S.: Focusing
  attention: Towards accurate text recognition in natural images.
\newblock In: 2017 IEEE International Conference on Computer Vision (ICCV), pp.
  5086--5094 (2017)

\bibitem{cheng2018aon}
{Cheng}, Z., {Xu}, Y., {Bai}, F., {Niu}, Y., {Pu}, S., {Zhou}, S.: Aon: Towards
  arbitrarily-oriented text recognition.
\newblock In: 2018 IEEE/CVF Conference on Computer Vision and Pattern
  Recognition, pp. 5571--5579 (2018)

\bibitem{Dalal2005Histograms}
Dalal, N., Triggs, B.: Histograms of oriented gradients for human detection.
\newblock In: IEEE Computer Society Conference on Computer Vision \& Pattern
  Recognition, pp. 886--893 (2005)

\bibitem{Deng2009ImageNet}
Deng, J., Dong, W., Socher, R., Li, L.J., Li, K., Li, F.F.: Imagenet: A
  large-scale hierarchical image database.
\newblock In: Computer Vision and Pattern Recognition, 2009. CVPR 2009. IEEE
  Conference on, pp. 248--255 (2009)

\bibitem{denton2015deep}
{Denton}, E.L., {Chintala}, S., {Szlam}, A., {Fergus}, R.: Deep generative
  image models using a laplacian pyramid of adversarial networks.
\newblock neural information processing systems pp. 1486--1494 (2015)

\bibitem{Goodfellow2014Generative}
Goodfellow, I.J., Pouget-Abadie, J., Mirza, M., Xu, B., Warde-Farley, D.,
  Ozair, S., Courville, A., Bengio, Y.: Generative adversarial networks.
\newblock Advances in Neural Information Processing Systems \textbf{3},
  2672--2680 (2014)

\bibitem{GTOCR}
Google: Tesseract optical character recognition.
\newblock \url{https://code.google.com/p/tesseract-ocr/} (2006)

\bibitem{GTF}
Google: Tensorflow.
\newblock \url{https://www.tensorflow.org/} (2016)

\bibitem{Ioffe2015Batch}
Ioffe, S., Szegedy, C.: Batch normalization: Accelerating deep network training
  by reducing internal covariate shift.
\newblock In: International Conference on Machine Learning, pp. 448--456 (2015)

\bibitem{Isola2017Image}
Isola, P., Zhu, J.Y., Zhou, T., Efros, A.A.: Image-to-image translation with
  conditional adversarial networks.
\newblock In: IEEE Conference on Computer Vision and Pattern Recognition, pp.
  5967--5976 (2017)

\bibitem{jaderberg2015deep}
{Jaderberg}, M., {Simonyan}, K., {Vedaldi}, A., {Zisserman}, A.: Deep
  structured output learning for unconstrained text recognition.
\newblock international conference on learning representations  (2015)

\bibitem{jaderberg2016reading}
{Jaderberg}, M., {Simonyan}, K., {Vedaldi}, A., {Zisserman}, A.: Reading text
  in the wild with convolutional neural networks.
\newblock International Journal of Computer Vision \textbf{116}(1), 1--20
  (2016)

\bibitem{Jaderberg2014Deep}
Jaderberg, M., Vedaldi, A., Zisserman, A.: Deep features for text spotting.
\newblock In: European Conference on Computer Vision, pp. 512--528 (2014)

\bibitem{karatzas2013icdar}
{Karatzas}, D., {Shafait}, F., {Uchida}, S., {Iwamura}, M., i~{Bigorda}, L.G.,
  {Mestre}, S.R., {Mas}, J., {Mota}, D.F., {Almazàn}, J.A., de~las {Heras},
  L.P.: Icdar 2013 robust reading competition.
\newblock In: 2013 12th International Conference on Document Analysis and
  Recognition, pp. 1484--1493 (2013)

\bibitem{Kingma2014Adam}
Kingma, D., Ba, J.: Adam: A method for stochastic optimization.
\newblock Computer Science  (2014)

\bibitem{lee2016recursive}
{Lee}, C.Y., {Osindero}, S.: Recursive recurrent nets with attention modeling
  for ocr in the wild.
\newblock In: 2016 IEEE Conference on Computer Vision and Pattern Recognition
  (CVPR), pp. 2231--2239 (2016)

\bibitem{Lucas2003ICDAR}
Lucas, S.M., Panaretos, A., Sosa, L., Tang, A., Wong, S., Young, R.: Icdar 2003
  robust reading competitions.
\newblock Proc of the Icdar \textbf{7}(2-3), 105--122 (2003)

\bibitem{Mishra2013Scene}
Mishra, A., Alahari, K., Jawahar, C.V.: Scene text recognition using higher
  order language priors (2013)

\bibitem{Radford2015Unsupervised}
Radford, A., Metz, L., Chintala, S.: Unsupervised representation learning with
  deep convolutional generative adversarial networks.
\newblock Computer Science  (2015)

\bibitem{Ronneberger2015U}
Ronneberger, O., Fischer, P., Brox, T.: U-net: Convolutional networks for
  biomedical image segmentation.
\newblock In: International Conference on Medical Image Computing and
  Computer-Assisted Intervention, pp. 234--241 (2015)

\bibitem{shi2017an}
{Shi}, B., {Bai}, X., {Yao}, C.: An end-to-end trainable neural network for
  image-based sequence recognition and its application to scene text
  recognition.
\newblock IEEE Transactions on Pattern Analysis and Machine Intelligence
  \textbf{39}(11), 2298--2304 (2017)

\bibitem{shi2018aster}
{Shi}, B., {Yang}, M., {Wang}, X., {Lyu}, P., {Yao}, C., {Bai}, X.: Aster: An
  attentional scene text recognizer with flexible rectification.
\newblock IEEE Transactions on Pattern Analysis and Machine Intelligence pp.
  1--1 (2018)

\bibitem{Simonyan2014Very}
Simonyan, K., Zisserman, A.: Very deep convolutional networks for large-scale
  image recognition.
\newblock Computer Science  (2014)

\bibitem{Tian2016Multilingual}
Tian, S., Bhattacharya, U., Lu, S., Su, B., Wang, Q., Wei, X., Lu, Y., Tan,
  C.L.: Multilingual scene character recognition with co-occurrence of
  histogram of oriented gradients.
\newblock Pattern Recognition \textbf{51}(C), 125--134 (2016)

\bibitem{Wang2013End}
Wang, T., Wu, D.J., Coates, A., Ng, A.Y.: End-to-end text recognition with
  convolutional neural networks.
\newblock In: International Conference on Pattern Recognition, pp. 3304--3308
  (2013)

\bibitem{Wang2017Learning}
Wang, Y., Shi, C., Xiao, B., Wang, C.: Learning spatially embedded
  discriminative part detectors for scene character recognition.
\newblock In: Iapr International Conference on Document Analysis and
  Recognition, pp. 363--368 (2017)

\bibitem{zeiler2014visualizing}
{Zeiler}, M.D., {Fergus}, R.: Visualizing and understanding convolutional
  networks.
\newblock european conference on computer vision pp. 818--833 (2014)

\bibitem{Zhang2018Robust}
Zhang, Y., Liang, S., Nie, S., Liu, W., Peng, S.: Robust offline handwritten
  character recognition through exploring writer-independent features under the
  guidance of printed data.
\newblock Pattern Recognition Letters \textbf{106} (2018)

\bibitem{zhang2016natural}
Zhang, Z., Xu, Y., Liu, C.L.: Natural scene character recognition using robust
  pca and sparse representation.
\newblock In: Document Analysis Systems (DAS), 2016 12th IAPR Workshop on, pp.
  340--345. IEEE (2016)

\end{thebibliography}

\end{document}